\documentclass[letterpaper, 10 pt, conference]{ieeeconf}  

\IEEEoverridecommandlockouts                              

\usepackage[dvipsnames]{xcolor}

\usepackage{graphicx} 
\usepackage{epsfig} 
\usepackage{mathptmx} 
\usepackage{times} 
\usepackage{amsmath,amssymb,amsfonts}
\usepackage{pgfplots}
\pgfplotsset{compat=1.7}
\usepackage{pgfplotstable}
\usepackage{tikz,pgfplots}
\usepackage{physics} 
\usepackage{siunitx} 
\usepackage{enumerate} 
\usepackage{tikz,pgfplots}
\usepackage{verbatim}
\usepackage{color}
\usepackage{mathtools}
\usepackage{breqn}
\usepackage{subcaption}
\usepackage{array}
\usepackage{makecell}
\usepackage{soul}  
\usepackage{comment}
\usepackage{textcomp}
\usepackage{subcaption}
\usepackage{algorithm}
\usepackage{algpseudocode} 
\usepackage{multirow}
\usepackage{multicol}
\usepackage{tabularx}

\newcommand{\jk}[1]{\textcolor{black}{#1}}

\usepackage[]{graphicx}

\def\BibTeX{{\rm B\kern-.05em{\sc i\kern-.025em b}\kern-.08em
    T\kern-.1667em\lower.7ex\hbox{E}\kern-.125emX}}

\newenvironment{conditions}
  {\par\vspace{\abovedisplayskip}\noindent\begin{tabular}{>{$}l<{$} @{${}={}$} l}}
  {\end{tabular}\par\vspace{\belowdisplayskip}}

\title{\LARGE \bf
Coordinated Motion Planning of a Wearable Multi-Limb System\\ for Enhanced Human-Robot Interaction

}

\author{Chaerim Moon and Joohyung Kim
\thanks{All the authors are with the KIMLAB (Kinetic Intelligent Machine LAB) at the University of Illinois, Urbana-Champaign, Urbana, IL 61801, USA. {\tt\small \{cm74, joohyung\}@illinois.edu.}}%
}

\begin{document}

\maketitle
\thispagestyle{empty}
\pagestyle{empty}

\begin{abstract}

Supernumerary Robotic Limbs (SRLs) can enhance human capability within close proximity. However, as a wearable device, the generated moment from its operation acts on the human body as an external torque. When the moments increase, more muscle units are activated for balancing, and it can result in reduced muscular null space. Therefore, this paper suggests a concept of a motion planning layer that reduces the generated moment for enhanced Human-Robot Interaction. It modifies given trajectories with desirable angular acceleration and position deviation limits. Its performance to reduce the moment is demonstrated through the simulation, which uses simplified human and robotic system models.
\end{abstract}

\section{Introduction}


SRLs provide an artificial limb to a human subject in addition to their natural limbs. Thus, their coordination with the natural limbs has been an important concern, and it introduced approaches such as utilizing the motor task null space - in terms of kinematic, muscular, and neural aspects \cite{dominijanni2021neural}. That is, muscle units that are not related to a task can be used as a controller of SRL. It implies that reducing the required number of muscle units for system operation will allow the human to possess more options to manipulate the robotic system.



Human motion control for static and dynamic balancing accompanies a coordinated body movement \cite{horak2006postural}. When there exists external force and moment applied to the human body, the human activates muscles and adjusts postures for balance \cite{wu2001control}. Increased external loads result in more muscle units engaged and greater joint angles changed. Thus, the motions of SRLs need to be designed considering the generated moments from their operation to ameliorate the physical burden on the human body.


 
Thus, in this paper, a motion planning layer to reduce the moments from a wearable multi-limb system operation is introduced. Its functional validity is examined under both static and dynamic initial conditions in the simulation environment.



\begin{figure}
    \centering
        \begin{subfigure}{0.47\columnwidth} 
		\includegraphics[width=0.99\columnwidth]{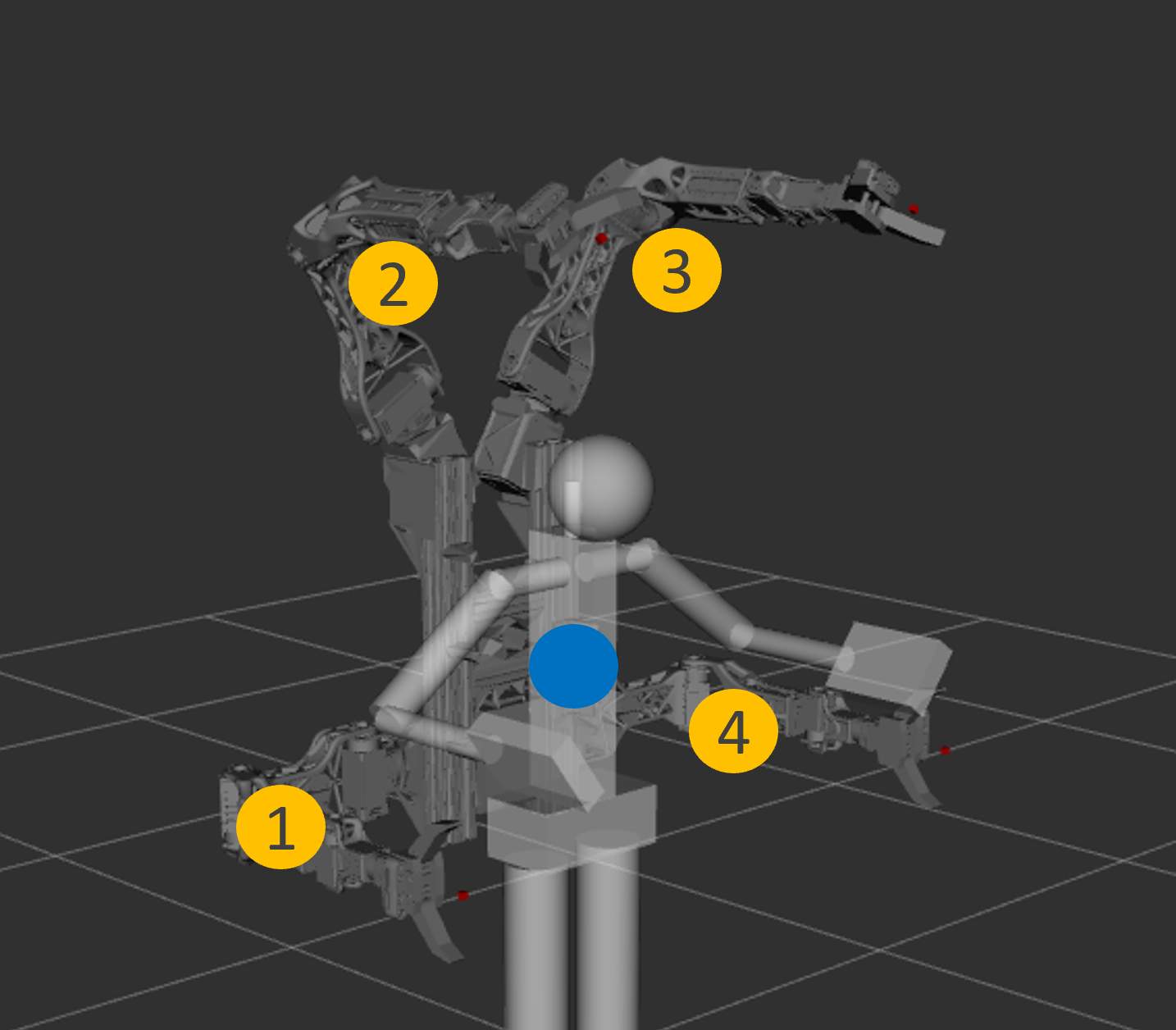}
		\caption{Neutral posture} 
	\end{subfigure}
	\begin{subfigure}{0.47\columnwidth} 
		\includegraphics[width=0.99\columnwidth]{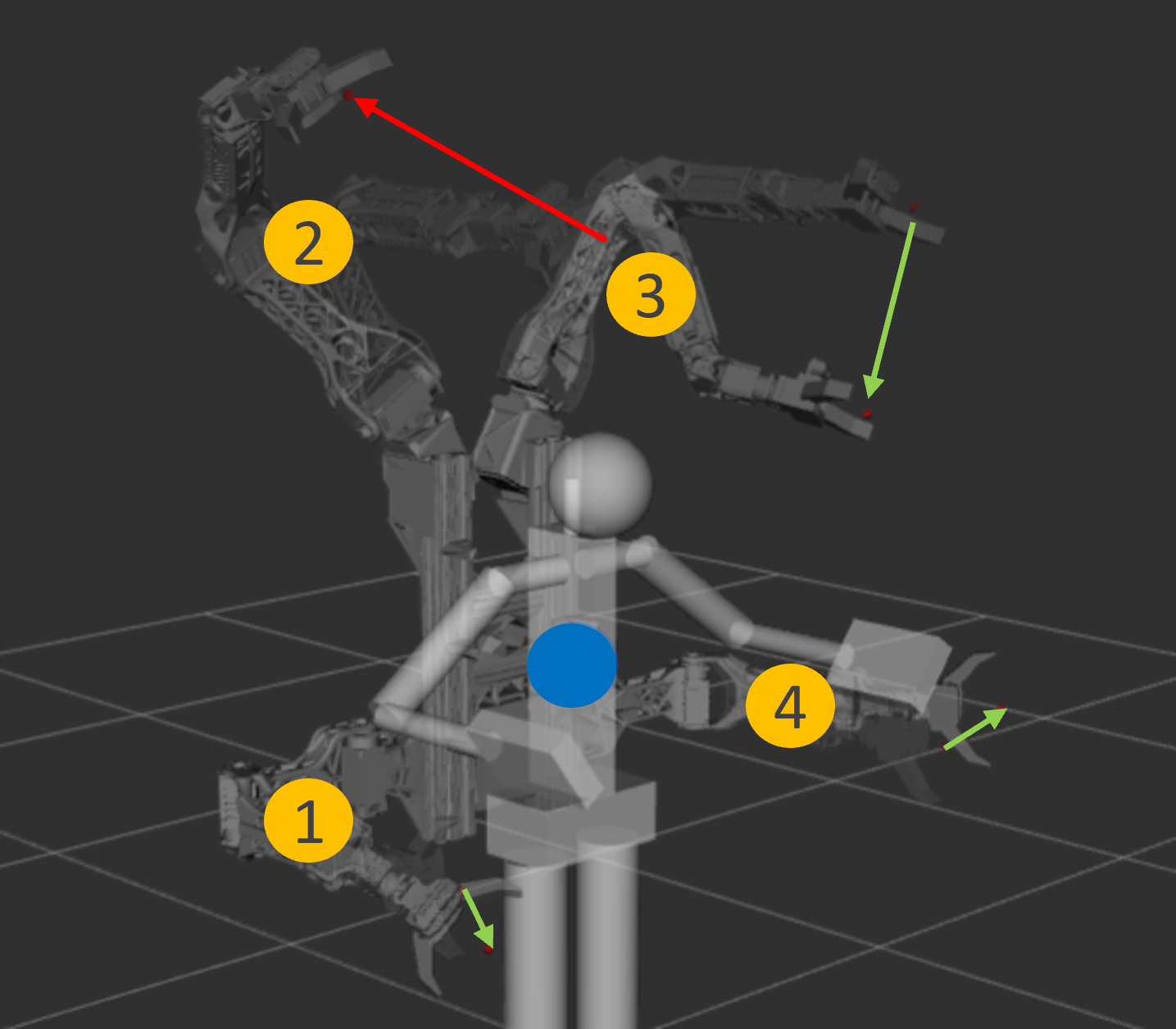}
		\caption{Unbalanced posture} 
	\end{subfigure}
    \caption{Visualization of a wearable multi-limb system in different configurations.}
    \vspace{-5mm}
    \label{fig:main}
\end{figure}

\section{System Dynamics Modelling}

Simplified models of a human subject and a wearable multi-limb system (Fig. \ref{fig:main}) are designed. A human subject is modelled considering the 50\textsuperscript{th} percentile male anthropometric data from \cite{kubicek1995man}. The reference point to calculate the generated moment $M_{human}$ is defined at the center of the thoracic vertebra T10 - the blue dot. Each robotic limb is assumed as a point mass - the yellow dots, and it rotates about a fixed coordinate described relatively to the human body frame.



Based on these models, the moment generated by the system operation can be expressed as follows:
\vspace{-3mm}

\begin{equation} \label{eq1}
\vec{M}_{human} = \sum_{i=1}^n \vec{r}_i \cross (m_i\vec{g} + m_i\vec{a}_{h,i})
\end{equation}
\vspace{-5mm}

\noindent where:

\begin{conditions}
 n     &  the number of robotic limbs \\
 \vec{r}_i    &  distance from a human body to $i_{th}$ limb's COM \\
 m_i     &  mass of $i_{th}$ robotic limb \\   
 \vec{g}     &  gravitational acceleration \\
 \vec{a}_{h,i}   &  acceleration of $i_{th}$ COM about a human body frame
\end{conditions}

The first term including $\vec{g}$ is related to the robotic system's center of mass (COM) deviation, and the term including $\vec{a}_h$ is related to the motion of the limbs.



\section{Multi-limb Trajectory Modification}
In order to reduce $M_{human}$, this study suggests a motion planning layer to modify current trajectories under certain circumstances. \jk{According to (\ref{eq1}), the total moment can be changed by adjusting $\vec{r}_i$ and $\vec{a}_{h,i}$. These variables are functions of each joint angular acceleration over time ($\vec{\alpha}_i(t)$), for the given initial joint states (i.e., joint angular position and velocity). In short, $M_{human}$ can be reduced by determining $\vec{\alpha}_i(t)$ for each robotic limb.}

The motion planning layer is activated when there exists a sudden robotic limb motion that increases the moment, and it modifies other limbs' trajectories with $\vec{\alpha}_i(t)$. Every desirable $\vec{\alpha}_i(t)$ is defined with two conditions: ({\romannumeral 1}) it minimizes $M_{human}$; ({\romannumeral 2}) the new trajectories deviate from the original ones within preset limits - it can be adjusted based on applications. When searching for $\vec{\alpha}_i(t)$, random combinations are selected, and the corresponding moment is calculated. The most desirable set is selected after 3,000 times of iterations for each control loop, and the trajectories are applied to each robotic limb.



\begin{table}
\centering
\begin{tabular}
{|c c c c c|}
\hline
Case \# & Initial states & \begin{math}\vec{\theta}_{prev,x}\end{math} (\textdegree) &
\begin{math}\vec{\omega}_{prev,x}\end{math} (\textdegree /s\textsuperscript{2}) & Comp.\\ [0.5ex]
\hline\hline
1-1 & zero & [0 0 0] & [0 0 0] & O\\
1-2 & & & & X\\[0.5ex]
\hline
2-1& non-zero & [40 -70 -20] & [10 -10 20] & O\\
2-2 & & & & X\\[0.5ex]
 \hline
\end{tabular}
\caption{Simulation conditions. The array of \begin{math}\vec{\theta}_{prev,x}\end{math} and \begin{math}\vec{\omega}_{prev,x}\end{math} are in the order of robotic limb \#1, \#3, and \#4. Comp. indicates the existence of moment compensation motions.}
    \vspace{-3mm}

\label{table:simulation}
\end{table}

\section{Simulation}

A simulation to evaluate the proposed motion modification was conducted using MATLAB R2022a. The mass of each robotic arm was assumed as 10 \% of a human body mass, and its length was assumed as the thumb-tip reach defined in the human model. In terms of motions, it was assumed that the robotic limb \#2 moves toward the backside of the human body with constant acceleration to represent the moment-increasing situation. Specifically, the motion was designed to displace its angle from $0^{\circ}$ to $90^{\circ}$ in 2.5 seconds. The $\vec{\alpha}_i(t)$ was selected in the range of $\pm20^{\circ}/s^2$, and the limit of angular displacement of compensating motions was confined as $\pm 20^{\circ}$ from their original trajectories. Under the assumptions, two conditions were considered - the initial states of the other arms are at zero states and at non-zero states (TABLE \ref{table:simulation}). The non-zero initial states were randomly selected within the range of the existing hardware specifications. The generated moments from each case were calculated and analysed.




\section{Result and Discussion}
\begin{figure}
    \centering
    \begin{subfigure}{0.99\columnwidth}
        \includegraphics[width=0.99\columnwidth]{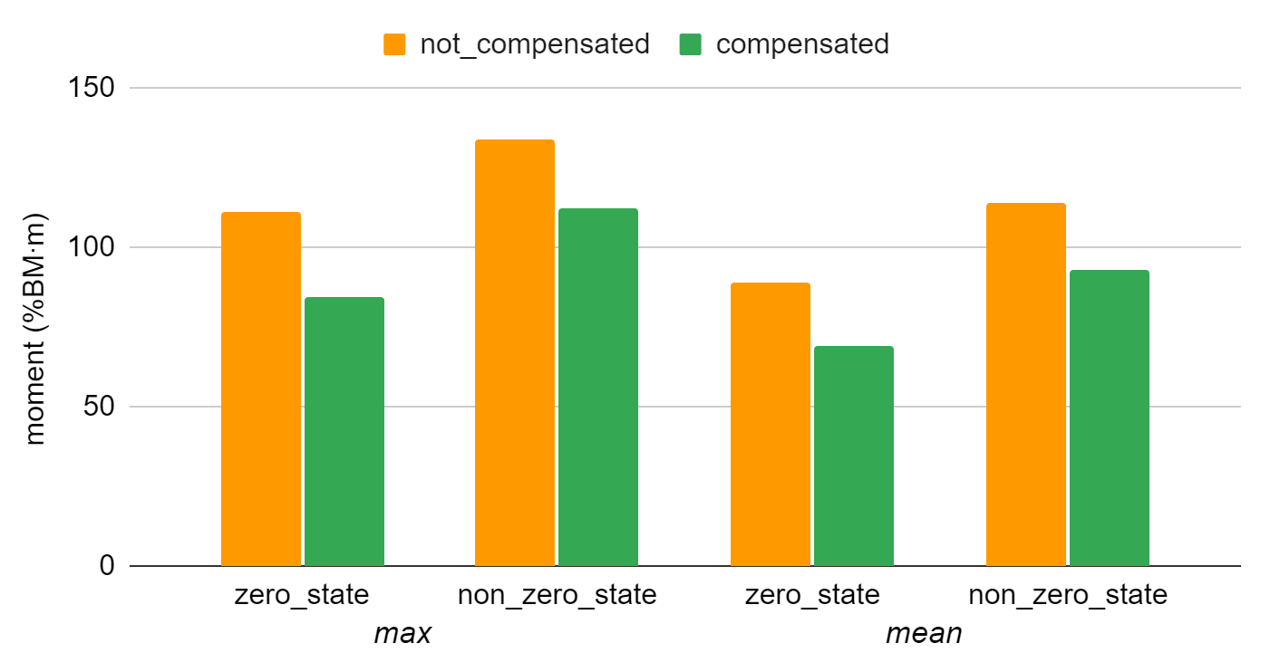}
		\caption{Max and mean moments with both initial conditions}
  \label{fig:moment_max_mean}
	\end{subfigure}
    \begin{subfigure}{0.99\columnwidth}
        \includegraphics[width=0.99\columnwidth]{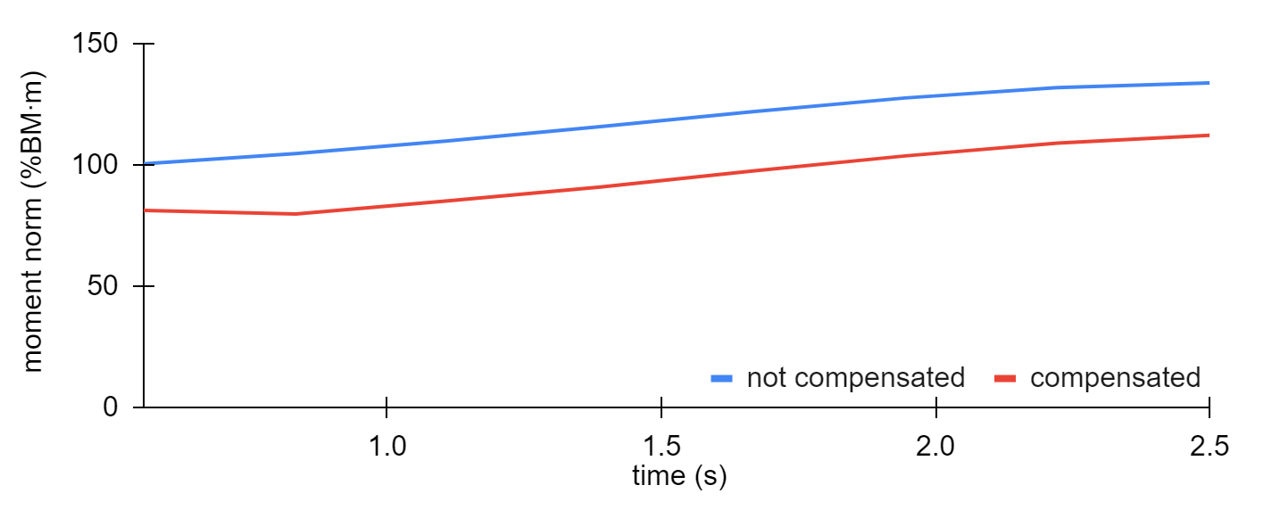}
		\caption{Moments with the non-zero initial condition}
  \label{fig:moment_profile}
	\end{subfigure}
    \caption{The results of simulation - the Euclidean norm of generated moments. \%BM indicates \%body mass.}
      \vspace{-3mm}
    \label{fig:result}
\end{figure}

The Euclidean norms of the generated moment were investigated under two initial state conditions (Fig. \ref{fig:result}). The result demonstrates that the moments were reduced by selecting desirable $\vec{\alpha}_i$s while not compromising the trajectory deviation limits. During the operation, the moments with both initial conditions decreased, indicating that its functional validity is regardless of the initial states - whether the system is at a static state, or it is following a certain trajectory.



When designing the robotic limb $\#$2 motion in the simulation, a collision avoidance situation was imitated. For example, when a robotic limb is about to collide with a human body segment, its trajectory needs to be modified. It includes cases moving to the backside of the human body as like the simulated motion, which increases $M_{human}$. It represents an example of target situations to activate the suggested motion planning layer. The simulation demonstrates how the trajectory modification reduced the moment.




The motion modification strategy was determined considering a control method for SRLs - utilizing muscular null space. Since $M_{human}$ acts on the human body as an external torque, which needs to be countered for balance, an objective function is set to reduce it. With less amount of torque to compensate, it is expected that a smaller number of muscle units will be engaged. It results in more units being secured to muscular null space, enlarging the available control modality.




A simplified robotic system model was used to concise the problem and concentrate on its main goal, which is to reduce the moment. In order to implement this motion modifier to real robotic applications such as the system depicted in Fig. \ref{fig:main}, the objective function will need to include the complicated system kinematics and dynamics. Additionally, it is necessary to consider the interaction between the robotic system and its surrounding environment.




\section{Conclusion}

This paper suggests a motion planning layer for a wearable multi-limb system. It is aimed to reduce the generated moment with respect to the human body. The simulation with simple human and robotic system models demonstrated its validity under static and dynamic initial conditions. It is expected that with this layer to modify robotic limb trajectories, the human will bear reduced physical burdens while securing greater muscular null space.


\addtolength{\textheight}{-12cm}   




\bibliographystyle{ieeetr}
\bibliography{reference}

\end{document}